\pdfoutput=1

\documentclass[11pt]{article}

\usepackage[final]{acl}

\usepackage{times}
\usepackage{latexsym}

\usepackage[T1]{fontenc}

\usepackage[utf8]{inputenc}

\usepackage{microtype}

\usepackage{inconsolata}

\usepackage{graphicx}
\usepackage{amsmath}
\usepackage{amsfonts}
\usepackage{booktabs}
\usepackage{algorithm}
\usepackage{algpseudocode}
\usepackage{caption}
\usepackage{tcolorbox}
\usepackage{caption}
\usepackage{subcaption}     
\usepackage{multirow}
\usepackage{array}    

\title{Enhancing Efficiency and Exploration in Reinforcement Learning for LLMs}

\author{
 \textbf{Mengqi Liao\textsuperscript{1,2}},
 \textbf{Xiangyu Xi\textsuperscript{2}},
 \textbf{Ruinian Chen\textsuperscript{2}},
 \textbf{Jia Leng\textsuperscript{2}},
\\
 \textbf{Yangen Hu\textsuperscript{2}},
 \textbf{Ke Zeng\textsuperscript{2}},
 \textbf{Shuai Liu\textsuperscript{1}},
 \textbf{Huaiyu Wan \textsuperscript{1,3}\footnote{Corresponding author}},
\\
\\
 \textsuperscript{1}School of Computer Science and Technology, Beijing Jiaotong University\\
 \textsuperscript{2}Meituan\\
  \textsuperscript{3}Beijing Key Laboratory of Traffic Data Mining and Embodied Intelligence
\\
 \small{
   \textbf{*Correspondence:} \href{mailto:hywan@bjtu.edu.cn}{hywan@bjtu.edu.cn}
 }
}

\begin{document}
\maketitle
\begin{abstract}
Reasoning large language models (LLMs) excel in complex tasks, which has drawn significant attention to reinforcement learning  (RL)  for LLMs. 
However, existing approaches allocate an equal number of rollouts to all questions during the RL process, which is  inefficient. This inefficiency stems from the fact that training on simple questions yields limited gains, whereas more rollouts are needed for challenging questions to sample correct answers.
Furthermore, while RL improves response precision, it limits the model's exploration ability, potentially resulting in a performance cap below that of the base model prior to RL.
To address these issues, we propose a mechanism for dynamically allocating rollout budgets based on the difficulty of the problems, enabling more efficient RL training. Additionally, we introduce an adaptive dynamic temperature adjustment strategy to maintain the entropy at a stable level, thereby encouraging sufficient exploration. This enables LLMs to improve response precision while preserving their exploratory ability to uncover potential correct pathways. The code and data is available on: \url{https://github.com/LiaoMengqi/E3-RL4LLMs}
\end{abstract}
\section{Introduction}

Large language models (LLMs) have gained considerable attention for their capabilities across a wide range of applications \cite{kumar2024large}. Recently, advanced reasoning models trained with reinforcement learning (RL)  , such as DeepSeek-R1 \cite{guo2025deepseek} and Kimi k1.5, \cite{team2025kimi} have demonstrated remarkable improvements in complex tasks like mathematics and coding, further intensifying research interest in RL for LLMs. After that, many works related to reinforcement learning for LLMs emerged \cite{xu2025towards,yu2025dapo,liu2025understanding}. Among them, the most common combination is to use the GRPO \cite{grpo} algorithm or its variants combined with rule-based rewards for reinforcement learning. 

However, rule-based rewards result in very sparse reward signals. When the training data is particularly challenging, the policy struggles to sample the correct answer, causing the advantage in the GRPO algorithm to become zero. In such cases, the policy fails to obtain an update gradient. DAPO \cite{yu2025dapo} filters out samples with a within-group reward standard deviation of zero to avoid zero advantage and employs multiple rounds of online sampling until enough experience is gathered for a single update. This approach is highly inefficient because the cost of sampling experiences is extremely high, yet this method results in a substantial waste of rollouts. 

What's more, \citet{yue2025does} highlights that during evaluation, while the RL model surpasses the base model with a small sample size (small \( k \)), the base model achieves superior pass@\( k \) performance as the sample size increases (large \( k \)). This occurs because reinforcement learning prioritizes maximizing rewards, leading the model to focus probabilities on high-reward paths, potentially neglecting diverse correct answers. 
While encouraging more exploration during training could potentially mitigate this issue, our experiments reveal that combining entropy regularization \cite{mnih2016asynchronous, entropy_loss}—the widely adopted approach for fostering exploration in deep reinforcement learning—with sparse rule-based rewards can degrade performance, especially when training with challenging questions, and may even result in model collapse.

To address the aforementioned issues, we first introduce a dynamic rollout budget allocation mechanism to enhance the training efficiency of RL. For simple questions that the model can answer proficiently, we reduce their rollout budget, as performing reinforcement learning on such problems yields minimal gains. The saved rollout budget is reallocated to more challenging problems, thereby increasing the likelihood of obtaining correct answers. 
Additionally, to promote exploration without introducing harmful gradients, we propose a temperature scheduler that dynamically adjusts the temperature to maintain a stable policy entropy level, thereby enabling more extensive exploration during training.  An annealing mechanism is further integrated to effectively balance exploration and exploitation. In summary, our contributions are as follows:  
\begin{itemize}
    \item We propose a dynamic rollout budget allocation mechanism that enables a more rational distribution of computational resources, allowing RL to be conducted more efficiently. 
    \item We introduce a temperature scheduler that dynamically adjusts the sampling distribution's temperature, maintaining the entropy at a stable level to encourage more exploration.
    \item  Experimental results demonstrate that our method improves the 7B model's pass@1 by 5.31\% and pass@16 by 3.33\% on the AIME 2024 benchmark compared to train with GRPO only, and consistently outperforms GRPO in pass@16 across various benchmarks.
\end{itemize}

\section{Related Work}
\textbf{Reinforcement Learning for LLMs.} \citet{ouyang2022training} trains a reward model using preference data and employs Proximal Policy Optimization (PPO) \cite{schulman2017proximal} to perform reinforcement learning on LLMs for alignment with human preferences. Subsequently, many new approaches have employed reinforcement learning to align LLMs with human preferences \cite{wang2024comprehensive}. DeepSeekMath \cite{grpo} proposed the GRPO algorithm, which simplifies the training process of reinforcement learning and significantly enhances the performance of LLMs in the mathematical domain through RL. Subsequently, DeepSeek-R1 \cite{guo2025deepseek} and Kimi k1.5 \cite{team2025kimi} successfully demonstrated the substantial impact of reinforcement learning combined with rule-based reward sets in enhancing the reasoning capabilities of models. \citet{liu2025understanding} and \citet{yu2025dapo}, among others, further introduced improvements to optimization algorithms to enhance training effectiveness.

\noindent \textbf{Exploration and Exploitation in Reinforcement Learning.} Balancing exploration and exploitation is a central challenge in reinforcement learning. Common strategies include \(\varepsilon\)-greedy, Upper Confidence Bounds (UCB), and Boltzmann Exploration 
 \cite{sutton1998reinforcement}. 
 Boltzmann Exploration selects actions based on a softmax probability distribution proportional to the exponential of the estimated values of actions, regulated by a temperature parameter \( \tau \). Similarly, LLMs generate tokens using a softmax distribution.
 \citet{asadi2017alternative} introduced the Mellowmax operator to enhance the stability of Softmax, while \citet{kim2019adaptive} applied meta-gradient reinforcement learning to dynamically adjust  temperature parameter of Mellowmax for better exploration-exploitation trade-offs. Moreover, Entropy Regularization, a common technique in deep reinforcement learning, adds an entropy term to the optimization objective to encourage stochastic policies and broader exploration \cite{entropy_loss}. Apart from these reinforcement learning-based methods, a self-improvement approach \citet{zeng2024b} similarly enhances training performance by maintaining the exploration capability of the LLM through adjustments among several discrete temperature levels.

\section{Preliminary}

\subsection{Group Relative Policy Optimization (GRPO)}

We utilize the GRPO \cite{grpo} algorithm to optimize the policy \(\pi_{\theta}\) (LLMs). GRPO estimates the advantage in a group-relative manner. Specifically, given a question-answer pair \((q, a) \sim \mathcal {D}\), the old policy \(\pi_{\theta_{\text{old}}}\) generates \(G\) individual responses \(\{o_i\}_{i=1}^G\) and then optimizes the policy model by maximizing the following objective:

\begin{align}
    \mathcal{J}_{\text{GRPO}}(\theta) = &\mathbb{E}_{(q, a) \sim \mathcal{D}, \{o_i\}_{i=1}^G \sim \pi_{\theta_{\text{old}}}(\cdot | q)} \nonumber \\
&\Bigg[
\frac{1}{G} \sum_{i=1}^G \frac{1}{|o_i|} \sum_{t=1}^{|o_i|}
\Big(
\min \Big( r_{i, t}(\theta) \hat{A}_{i, t}, \nonumber\\
&\, \text{clip}(r_{i, t}(\theta), 1 - \epsilon, 1 + \epsilon) \hat{A}_{i, t} \Big) \nonumber\\
& - \beta D_{\text{KL}}(\pi_\theta \| \pi_{\text{ref}})
\Big)
\Bigg],
\end{align}
where $ r_{i, t}(\theta) = \frac{\pi_\theta(o_{i, t} \mid q, o_i, <_t)}{\pi_{\theta_{\text{old}}}(o_{i, t} \mid q, o_i, <_t)}
$ and the advantage $\hat{A}_{i, t}$ is computed as:

\begin{equation}
\hat{A}_{i, t} = \frac{r_i - \text{mean}(\{r_j\}_{j=1}^G)}{\text{std}(\{r_j\}_{j=1}^G)}.   
\label{eq:adv}
\end{equation}

Here, \(r_i\) is the reward of response $o_i$. The term \(D_{\text{KL}}(\pi_\theta \| \pi_{\text{ref}})\) represents the KL divergence penalty, which is used to prevent the policy from deviating excessively from the initial policy \(\pi_{\text{ref}}\). Following prior work \cite{yu2025dapo,liu2025understanding}, we do not employ this penalty term.  Therefore, we do not elaborate further on this detail. 

\section{Methodology}

In this section, we first introduce our method for modeling question difficulty and propose a dynamic rollout budget allocation mechanism to allocate budgets based on question difficulty, improving training efficiency and the model's ability to answer complex questions. Next, we introduce a temperature scheduler to maintain the policy entropy, enhancing exploration, and further combine it with an annealing mechanism to balance exploration and exploitation.

\subsection{Dynamic Rollout Budget Allocation}

More challenging questions require a greater number of samples to obtain the correct answer. To allocate computational resources more efficiently, we transfer the rollout budget from simpler questions to more difficult ones, thereby enhancing the model's ability to address challenging problems.

We first introduce the method for modeling question difficulty. The RL training dataset \( \mathcal{D} = \{ (q_1, a_1), (q_2, a_2), \ldots, (q_{|\mathcal D |}, a_{|\mathcal D |}) \} \) consists of questions \( q_i \), their corresponding ground truth answers \( a_i \). For each question \( q_i \), its cumulative rollout count \( n_i^c \) and cumulative reward \( r_i^c \) are recorded.
At the end of each dataset iteration, data points are ranked by their average reward \( \frac{r_i^c}{n_i^c} \). The descending order of \( q_i \)'s average reward is \( \text{rank}(q_i) \), and its normalized ranking is \( k_i = \frac{\text{rank}(q_i)}{|\mathcal{D}|} \), where a larger \( k_i \) indicates higher difficulty.

\begin{algorithm}[t]
    \caption{Dynamic Rollout Budget}
    \label{alg:dynamic_rollout_simple}
    \begin{algorithmic}[1]
        \Require A batch of rankings $\{k_{(1)}, \dots, k_{(B)}\}$, $G$, $G_{\text{max}}$, $G_{\text{min}}$
        \State Total rollout budget $N_{\text{total}} = B \times G$
        \State For $i$ in $\{1,2,\dots,B\}$, initialize $G_{(i)} = G_{\text{min}}$
        \State Remaining rollouts budget $N_{\text{rem}} = N_{\text{total}} - B \times G_{\text{min}}$
        \State For $i$ in $\{1,2,\dots,B\}$, $G_{(i)} = G_{(i)} + \lfloor N_{\text{rem}} \times \frac{k_{(i)}}{\sum_{j=1}^B k_{(j)}} \rfloor$
        \State $N_{\text{rem}} = N_{\text{total}} - \sum_{i=1}^B G_{(i)}$
        \State Distribute $N_{\text{rem}}$ greedily based on descending order of $k_{i}$, respecting $G_{\text{max}}$ for each $G_{(i)}$
        \State \Return $\{G_{(1)}, \dots, G_{(B)}\}$
    \end{algorithmic}
\end{algorithm}



After defining the difficulty of the questions, we allocate the rollout budget based on the identified difficulty levels. Specifically, we define the default, minimum, and maximum sampling budgets as \( G \), \( G_{\text{min}} \), and \( G_{\text{max}} \), respectively. The sampling budget \( G_i \) for question \( q_i \) is determined based on \( k_i \), such that larger \( k_i \) values correspond to higher allocated rollout budgets. The dynamic sampling budget allocation process is detailed in Algorithm~\ref{alg:dynamic_rollout_simple}, which ensures that the total rollout budget within a batch remains constant.

To avoid inefficiencies in allocating higher budgets to challenging questions under an undertrained policy, \( G_{\text{min}} \) and \( G_{\text{max}} \) are initially set equal to \( G \). After each iteration of \( \mathcal{D} \), \( G_{\text{max}} \) is gradually increased, and \( G_{\text{min}} \) is progressively decreased until reaching predefined limits. This prevents overcommitting resources to difficult questions prematurely and is analogous to curriculum learning.

\subsection{Temperature Scheduling to Promote Exploration}

\begin{figure*}[t]
  \centering
  \includegraphics[width=0.9\textwidth]{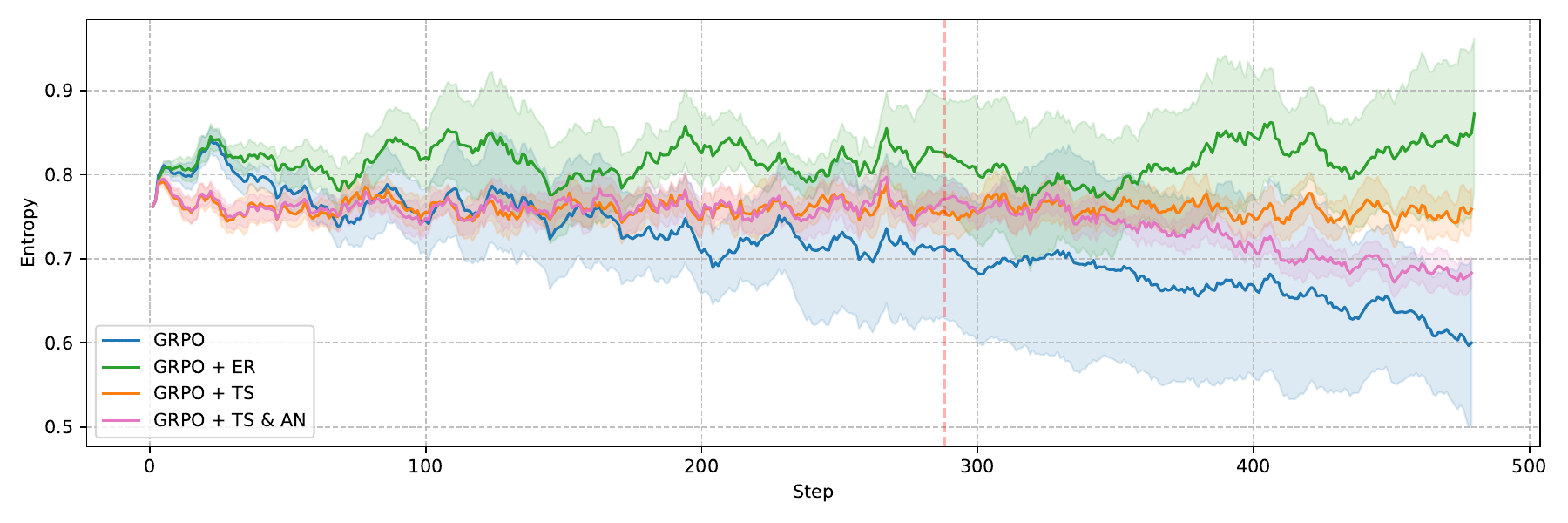}
  \caption{Smoothed entropy variations during training under different configurations. The curves represent the mean values, while the shaded regions denote the standard deviation across multiple runs. Here, \textbf{ER} represents entropy regularization, \textbf{TS} refers to the temperature scheduler, and \textbf{AN} indicates annealing. The red vertical line indicates the step at which annealing begins.}
  \label{fig:entropy}
\end{figure*}

\begin{figure*}[t] %
    \centering
    \includegraphics[width=0.7\linewidth]{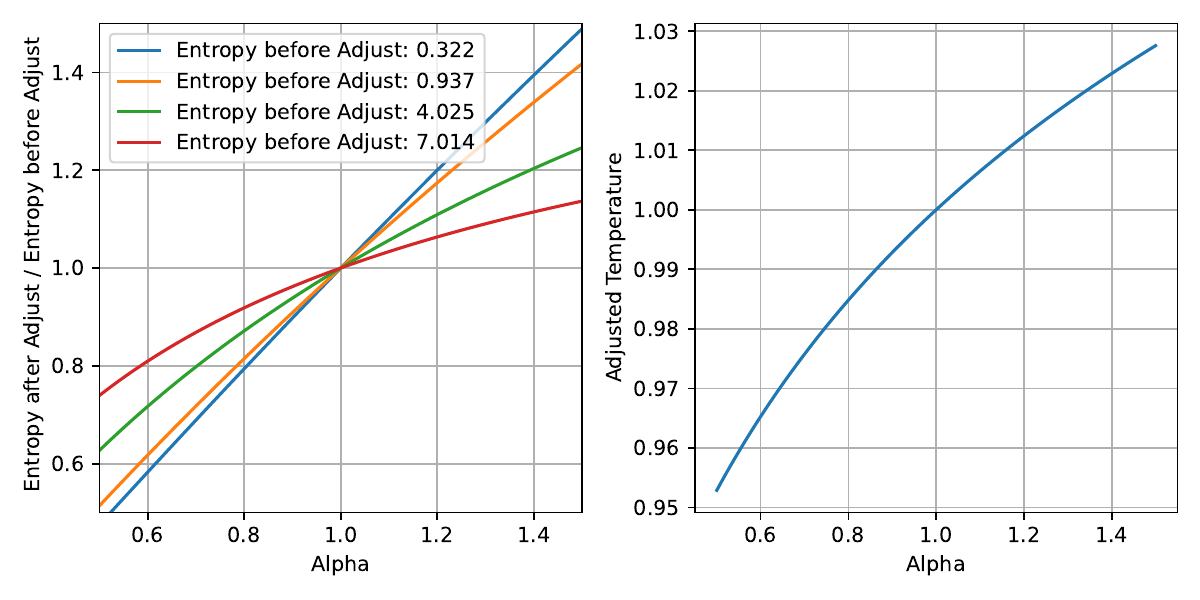} 
    \caption{The left figure illustrates the relationship between the scaling factor of $H_{t}$, after temperature adjustment, and $\alpha$. When the entropy is relatively small (the entropy magnitude  of distribution for next token generation is typically on the order of $10^{-1}$), the scaling factor closely approximates a linear relationship with $\alpha$. The right figure illustrates the relationship between $\tau_{t+1}$ and $\alpha$ when $\tau_t = 1$.}
    \label{fig:tpsc}
\end{figure*}

In reinforcement learning, policies may converge to local optima, hindering the discovery of the global optimum. By adding an entropy regularization term to the optimization objective, the strategy can be encouraged to  explore more state and action \citep{mnih2016asynchronous, entropy_loss}. The modified optimization objective is given by:  
\begin{align}
    \mathcal J(\theta)&=\mathcal J_\text{GRPO}(\theta) \nonumber\\&+\lambda \;\mathbb{E}_{(q, a) \sim \mathcal{D}, \{o_i\}_{i=1}^G \sim \pi_{\theta_{\text{old}}}(\cdot | q)}H(\pi_{\theta}(o_i|q)),
\end{align}
where \( H(\pi_{\theta}(o_i|q)) \) represents the Shannon entropy (Equation~(\ref{eq:entropy}) for the specific definition) of the action (token) sampling distribution from the policy, and \( \lambda \) is the coefficient controlling the strength of the regularization term. Entropy measures the uncertainty of a distribution, providing an indication of the policy's level of exploration.
However, rule-based rewards are very sparse. When the rewards  of all rollouts for a question are identical, the advantage $\hat A_{i,t}$ is zero, resulting in the gradient of the GRPO optimization objective, \(\nabla_\theta J_\text{GRPO}(\theta)\), is zero. In such cases, the update of the policy is primarily influenced by the gradient of the entropy regularization term, \(\nabla_\theta H(\pi_\theta)\). As training progresses, cases where the advantage equals zero become more frequent (as the proportion of fully correct rollouts increases), at which point the gradient of entropy regularization may potentially lead to the gradual collapse of the policy.

Figure~\ref{fig:entropy} shows the entropy evolution under different training setups. With GRPO alone, the policy's entropy declines rapidly, reducing the exploration of policy. The incorporation of entropy regularization effectively sustains higher entropy levels, thus fostering more diverse policy exploration.
However, the experimental results in Section \ref{sct:exp_ts} show that the performance of the model trained with entropy regularization is even worse than that of the model trained with GRPO alone.

\noindent \textbf{Temperature Scheduler.} As discussed, although entropy regularization helps maintain the entropy of the policy, it may inadvertently introduce harmful gradients. To address this, we propose a temperature scheduler that adaptively adjusts the temperature $\tau$ of the softmax distribution to maintain policy entropy, ensuring stable exploration without introducing additional gradients.
We aim to control the scaling of entropy by adjusting the temperature to maintain entropy at a stable level. However, the relationship between entropy and temperature is not linear. Fortunately, the entropy of the distributions for next token generation are typically small. Under this premise, we can adjust the temperature to precisely control entropy scaling using the following formula:
\begin{equation}  
\tau_{t+1} = \tau_{t} \times \left(1 + \frac{\tau_{t} \ln \alpha}{\ln |\mathcal{V}| + \ln\left(\ln |\mathcal{V}|\right)}\right), 
\label{eq:tp_adj}
\end{equation}  
where $\alpha = \frac{H_\text{init}}{H_\text{t}}$, with $H_\text{init}$ representing the average entropy of the first batch, which is the desired entropy level to maintain, and $H_\text{t}$ denoting the average entropy at the current training step $t$. Additionally, $|\mathcal{V}|$ represents the vocabulary size of the LLM.

Formula (\ref{eq:tp_adj}) ensures that the scaling factor of entropy, after temperature adjustment, maintains an approximately linear relationship with $\alpha$, as illustrated in Figure \ref{fig:tpsc}. \textbf{This indicates that entropy returns to the level of $H_\text{init}$ after the  temperature  adjustment.} The detailed derivation of Formula (\ref{eq:tp_adj}) is provided in Appendix \ref{sct:entropy_scaling}.
The temperature scheduler maintains the policy's entropy consistently at a stable level, as shown in Figure~\ref{fig:entropy}, thereby effectively enhancing policy exploration. Moreover, it is important to note that logits are also divided by the temperature during forward propagation after sampling, ensuring consistency between the LLM's distribution during training and sampling.

\noindent \textbf{Annealing Mechanism.} In the early stages of training, the policy requires sufficient exploration to avoid premature convergence to suboptimal solutions. As training progresses, we expect the policy to increasingly focus on exploiting high-value actions, thereby optimizing its performance more effectively. To balance exploration and exploitation, we introduce an annealing mechanism. Once the training step \( t \geq t_{\text{anneal}} \), \( \alpha = \frac{H_{\text{anneal}}^{(t)}}{H_{\text{t}}} \), where \( H_{\text{anneal}}^{(t)} \) is calculated as follows:
\begin{align}
    H_{\text{anneal}}^{(t)} &= H_{\text{init}} \cdot \left[ \eta + (1-\eta) \cdot \frac{1}{2}\left(1 + \right.\right.\nonumber\\
    & \left. \cos(\pi \cdot \frac{t - t_{\text{anneal}}}{t_{\max} - t_{\text{anneal}}})) \right]    .
\end{align}

Here, \( t_{\text{max}} \) is the maximum training steps, \( \eta \in [0,1) \).  Through this formula, We can gradually reduce the expect entropy level from \( H_{\text{init}} \) to \( \eta \cdot H_{\text{init}} \). As shown in Figure~\ref{fig:entropy}, by introducing annealing, the entropy of the policy gradually decreases during the annealing phase. This facilitates a smooth transition of the policy from a high-entropy exploratory state to a lower-entropy exploitative state, ensuring that the policy maintains sufficient exploration during the early stages of training while gradually becoming more focused and efficient as training progresses.

\section{Experiments}
\subsection{Setting}

\textbf{Training Datasets and Benchmarks.} We follow DeepScaleR \cite{deepscaler2025} in selecting MATH \cite{hendrycks2measuring}, AIME 1983-2023 \cite{aops_aime}, Omni-MATH \cite{gaoomni}, and AMC (prior to 2023) as our training datasets. We follow Kimi K1.5 \cite{team2025kimi} to enhance RL training efficiency by balancing the difficulty of the questions. In total, we collected 10k high-quality data points as the training set and 0.5k data points as the validation set. Further details are provided in Appendix \ref{sct:dataset}. We evaluate on AIME 2024, AMC 2023, MATH 500 \cite{hendrycks2measuring}, and OlympiadBench \cite{he2024olympiadbench}.

\noindent \textbf{Training Details.} During sampling, the batch size is 64, with the default number of rollouts per question (\( G \)) set to 8. The sampling temperature is 1, and the maximum response length is 6k. Training is performed over 3 epochs on the 10k dataset, totaling 480 steps. We use DeepSeek-R1-Distill-Qwen 1.5B and 7B \cite{guo2025deepseek} as base models. For the 1.5B model, the learning rate is \( 5 \times 10^{-6} \), and for the 7B model, it is \( 2 \times 10^{-6} \). The policy update batch size is \( 64 \times 8 \), and experiences from each sampling are used to update the policy only once. 
For the 7B model, training was conducted on 8 NVIDIA A100 GPUs, requiring approximately $8 \times 36$ GPU hours per experiment. For the 1.5B model, training was performed on 4 NVIDIA A100 GPUs, taking approximately $4 \times 24$ GPU hours per experiment. To ensure the reliability of results given the randomness in RL, each experiment was repeated 3 times. The training code is adapted from the VeRL framework \cite{sheng2024hybridflow}.

\noindent \textbf{Evaluation Protocol.} Unless otherwise specified, for each question, we default to sampling 16 times under the temperature of 1, with a maximum response length of 6k tokens. We use pass@1 and pass@16 \cite{chen2021evaluating} as our evaluation metric. \textbf{We report pass@16 because it reflects the model's potential to explore more solution paths to solve the questions.} The average metrics across the 3 runs are reported. 

\subsection{Main Results}
\label{sct:baseline}

\begin{table*}[!t]
\centering
\resizebox{\textwidth}{!}{%
\begin{tabular}{llrrrrrrrrrr}
\toprule
\multirow{2}{*}{Size} & \multirow{2}{*}{Method} & \multicolumn{2}{c}{AIME 2024} & \multicolumn{2}{c}{AMC 2023} & \multicolumn{2}{c}{MATH 500} & \multicolumn{2}{c}{Olympiad-Bench}&\multicolumn{2}{c}{Average} \\
\cmidrule(lr){3-4} \cmidrule(lr){5-6} \cmidrule(lr){7-8} \cmidrule(lr){9-10} \cmidrule(lr){11-12}
 & & Pass@1 & Pass@16 & Pass@1 & Pass@16 & Pass@1 & Pass@16 & Pass@1 & Pass@16  & Pass@1 & Pass@16 \\
\midrule
\multirow{3}{*}{7B} & GRPO & 37.5 & 73.33 & 70.06 & 91.56 & 80.32 & 97.8 & \textbf{53.72} & 79.85 & 60.40 & 85.63\\
 & DAPO & 36.87 & 70.0 & 67.77 & \textbf{95.18} & 77.63 & 97.39 & 50.50 & 80.88 &58.19 & 85.86\\
 & Ours & \textbf{42.81} & \textbf{76.66} & \textbf{70.20} & 93.57 & \textbf{81.09} & \textbf{98.33} & 53.70 & \textbf{82.02}& \textbf{61.95} & \textbf{87.64} \\
\midrule
\multirow{3}{*}{1.5B} & GRPO & 24.66 & 59.76 & 60.73 & 88.79 & \textbf{72.97} & 95.86 & 46.33 & 74.66 &51.17&79.76\\
 & DAPO & 19.16 & 60.0 & 52.86 & 84.33 & 68.81 & 94.6 & 41.17 & 70.81&45.50&77.43 \\
 & Ours & \textbf{27.70} & \textbf{66.66} & \textbf{62.95} & \textbf{90.36} & 72.88 & \textbf{96.39} & \textbf{46.35} & \textbf{75.40}&\textbf{52.47}&\textbf{82.20} \\
\bottomrule
\end{tabular}%
}
\caption{Baseline comparison across different benchmarks.}
\label{tab:baseline}
\end{table*}

\begin{figure*}[!ht]
    \centering 
    \begin{minipage}[!t]{0.45\textwidth}
        \begin{minipage}[!t]{\textwidth}
        \centering 
        \includegraphics[width=\linewidth]{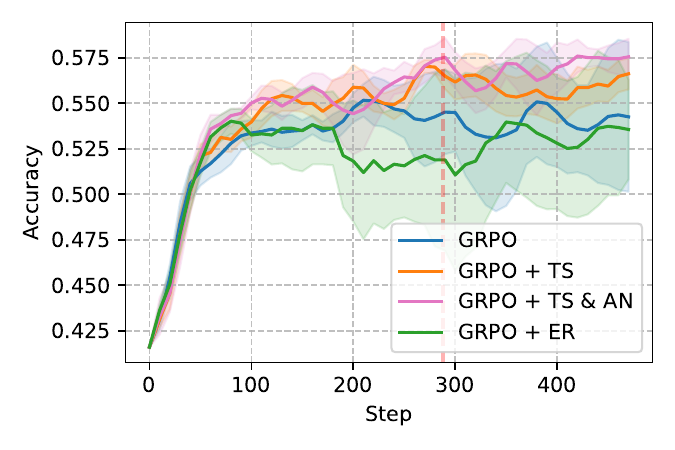}
        \vspace{-5pt} 
        \caption{The accuracy on the validation set during training, with the shaded area representing the variance across multiple runs. The red vertical line indicates the step at which annealing begins.}
        \label{fig:ts_acc}
    \end{minipage}
    \end{minipage}
    \hfill 
    \begin{minipage}[!t]{0.45\textwidth}
        \centering 
        \includegraphics[width=\linewidth]{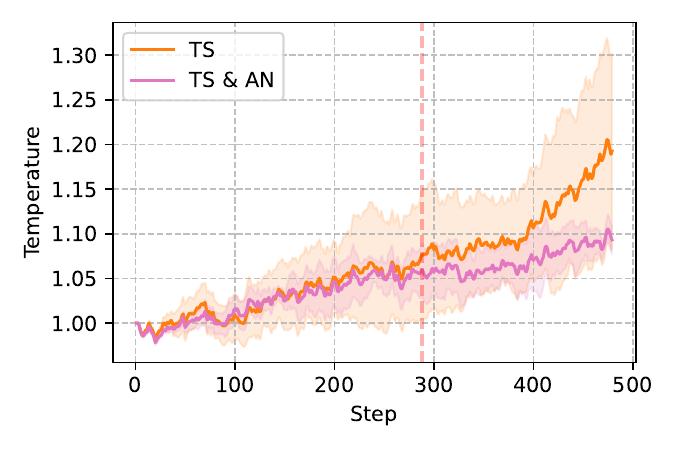}
        \vspace{-5pt} 
        \caption{The temperature variation during training is presented for cases utilizing only the temperature scheduler and for those combining the scheduler with annealing.}
        \label{fig:ts_tp}
    \end{minipage}
\end{figure*}

In this section, we compare our approach, which integrates dynamic rollout budget allocation, temperature scheduling, and annealing, with the baselines. 

\noindent \textbf{Baselines.} We select GRPO \cite{grpo} and DAPO \cite{yu2025dapo} as the baselines. Our proposed method is based on GRPO, which justifies its selection as a baseline. DAPO is a modified variant of GRPO, which filters out rollouts with zero advantage  and obtains experience through multiple rounds of online sampling. The general parameters for GRPO and DAPO are kept consistent with our method, while the DAPO-specific parameters are set to their default values as specified in its original paper.

\noindent \textbf{Implementation Details.}  For dynamic rollout budget allocation (\textbf{DR}), $G_{\text{max}}$ is increased by 2 and $G_{\text{min}}$ is decreased by 2 after each epoch. 
For annealing (\textbf{AN}), we test $\eta \in \{0.8, 0.85, 0.9\}$, observing instability for $\eta=0.8$ and $\eta=0.85$. Consequently, $\eta$ is set to 0.9, with the annealing start step at $\lfloor 0.6 \times t_{\text{max}} \rfloor$. 

\noindent \textbf{Analysis.} As shown in Table \ref{tab:baseline}, for the 7B model, our method achieves advantages of 5.31\% and 3.33\% in pass@1 and pass@16, respectively, on the AIME benchmark. On other benchmarks, our method also demonstrates significantly higher pass@16 performance compared to GRPO, indicating that the models trained with our method possess greater exploratory potential. The models trained with our method also achieve the best average pass@1 and pass@16 across the four benchmarks.
For the 1.5B model, our method exhibits similar advantages, demonstrating a significant improvement on the AIME benchmark and achieving the best average pass@1 and pass@16. The 1.5B model trained with DAPO performs significantly worse than those trained with other methods. This may be due to the 1.5B model's poor performance at the early stages of training, requiring numerous sampling rounds to gather enaugh experience for a single update. This results in substantial data being discarded, leading to its inferior performance. In contrast, GAPO and our method allow for more frequent policy updates, enabling faster improvement in model performance and more effective utilization of training data.

\subsection{The Impact of the Temperature Scheduler and Annealing}
\label{sct:exp_ts}

\begin{figure*}[!t]
    \centering 
    \begin{minipage}[!t]{0.45\textwidth}
        \begin{minipage}[!t]{\textwidth}
        \centering 
        \includegraphics[width=\linewidth]{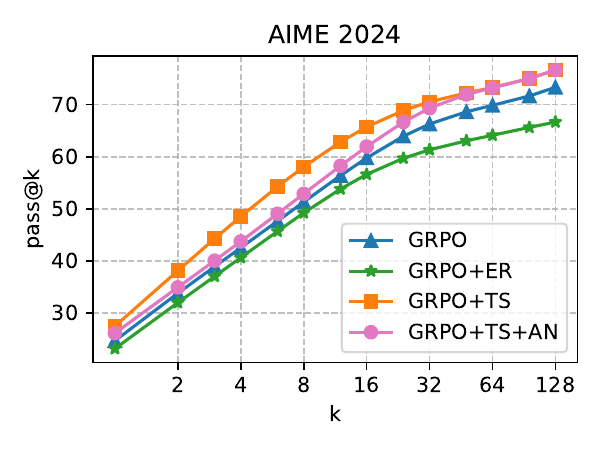}
        \vspace{-5pt} 
        \caption{Pass@k on AIME 2024}
        \label{fig:aime_passk}
    \end{minipage}
    \end{minipage}
    \hfill 
    \begin{minipage}[!t]{0.45\textwidth}
        \centering 
        \includegraphics[width=\linewidth]{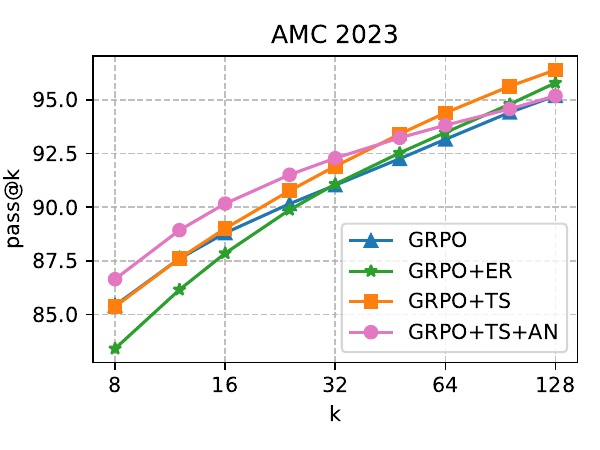}
        \vspace{-5pt} 
        \caption{Pass@k on AMC 2023}
        \label{fig:amc_passk}
    \end{minipage}
\end{figure*}

\begin{table*}[t]
\centering
\resizebox{\textwidth}{!}{
\begin{tabular}{lcccccccccc}
\toprule
\textbf{Method} & \multicolumn{2}{c}{\textbf{AIME 2024}} & \multicolumn{2}{c}{\textbf{AMC 2023}} & \multicolumn{2}{c}{\textbf{MATH 500}} & \multicolumn{2}{c}{\textbf{Olympiad-Bench}}& \multicolumn{2}{c}{\textbf{Average}} \\
\cmidrule(lr){2-3} \cmidrule(lr){4-5} \cmidrule(lr){6-7} \cmidrule(lr){8-9}\cmidrule(lr){10-11}
               & pass@1 & pass@16 & pass@1 & pass@16 & pass@1 & pass@16 & pass@1 & pass@16& pass@1 & pass@16 \\
\midrule
GRPO          & 24.66  & 59.76   & \textbf{60.73}  & 88.79   & 72.97  & 95.86   & 46.33  & 74.66  & 51.17 & 79.76\\
GRPO+ER       & 23.15  & 56.66   & 57.31  & 87.85 & 72.50  & 95.80   & 45.73  & 73.18  &49.67 & 78.37 \\
GRPO+TS       & \textbf{27.15} & \textbf{65.55} & 59.11 & 89.00 & 73.30  & \textbf{96.73} & 46.16  & \textbf{76.04} & 51.43 & \textbf{81.83}\\
GRPO+TS+AN    & 26.11  & 61.95   & 59.91  & \textbf{90.16}   & \textbf{74.66} & 96.20   & \textbf{46.78} & 74.41 & \textbf{51.86}&80.68 \\
\bottomrule
\end{tabular}
}
\caption{Comparison of different training methods on various benchmarks.}
\label{tab:ts_cmp}
\end{table*}

In this section, we analyze the impact of the temperature scheduler (\textbf{TS}) on the training of reinforcement learning, and compare it to entropy regularization (\textbf{ER}). For entropy regularization, $\lambda$ is set to $1 \times 10^{-4}$. For smaller benchmarks (AIME 2024 and AMC 2023), 128 answers per problem are sampled. For larger benchmarks (MATH 500 and Olympiad-Bench), the default sampling size of 16 is used.

\noindent  \textbf{Temperature Scheduler Maintains Entropy at a Stable Level.} Figure \ref{fig:entropy} illustrates the entropy variation under different configurations during training. As discussed earlier, solely using GRPO results in a rapid entropy decline, leading to insufficient exploration. In contrast, training with a temperature scheduler maintains entropy at a stable level, facilitating greater exploration. The introduction of annealing gradually reduces the entropy of the policy during the annealing phase, enabling a transition from an exploratory state to a more efficient exploitation state. Figure \ref{fig:ts_tp} illustrates the temperature variation, where annealing slows the temperature increase during the annealing phase.

\noindent \textbf{Temperature Scheduler Stabilizes LLM Performance Improvements.} Figure \ref{fig:ts_acc} shows the variation in validation accuracy during training. Both GRPO alone and GRPO with entropy regularization exhibit significant variance. \textbf{In contrast, training with the temperature scheduler achieves lower variance and higher accuracy, indicating that temperature scheduling enhances training stability and effectiveness.} This is because increased exploration prevents the policy from becoming trapped in local optima, resulting in more stable performance improvements.

\noindent \textbf{The Impact of the Temperature Scheduler on Performance.} 
As shown in Figure~\ref{fig:aime_passk} and Figure~\ref{fig:amc_passk}, models trained with the temperature scheduler generally outperform GRPO-only baselines in pass@k metrics, with the performance advantage consistently increasing as \(k\) grows. The experiments presented in Table~\ref{tab:ts_cmp} further demonstrate that incorporating the temperature scheduler significantly improves pass@16 compared to training solely with GRPO. In contrast, models trained with entropy regularization typically underperform relative to other methods, showing a slight advantage over GRPO-only training only when \(k\) is very large on the AMC benchmark.

\begin{table*}[!h]
\centering
\resizebox{\textwidth}{!}{%
\begin{tabular}{llrrrrrrrrrr}
\toprule
\multirow{2}{*}{Size} & \multirow{2}{*}{Method} & \multicolumn{2}{c}{AIME 2024} & \multicolumn{2}{c}{AMC 2023} & \multicolumn{2}{c}{MATH 500} & \multicolumn{2}{c}{Olympiad-Bench} &\multicolumn{2}{c}{Average} \\
\cmidrule(lr){3-4} \cmidrule(lr){5-6} \cmidrule(lr){7-8} \cmidrule(lr){9-10} \cmidrule(lr){11-12}
 & & Pass@1 & Pass@16 & Pass@1 & Pass@16 & Pass@1 & Pass@16 & Pass@1 & Pass@16 &Pass@1 & Pass@16 \\
\midrule
\multirow{2}{*}{7B} & All & \textbf{42.81} & \textbf{76.66} & 70.20 & \textbf{93.57} & 81.09 & \textbf{98.33} & \textbf{53.70} & \textbf{82.02} & \textbf{61.95} & \textbf{87.64} \\
 & w/o DS & 39.79 & 73.33 & \textbf{70.48} & 91.96 & \textbf{81.15} & \textbf{98.33} & 52.81 & 80.59 & 61.05 & 86.05\\
\midrule
\multirow{2}{*}{1.5B} & All & \textbf{27.70} & \textbf{66.66} & \textbf{62.95} & \textbf{90.36} & 72.88 & \textbf{96.39} & 46.35 & \textbf{75.40} & \textbf{52.47} & \textbf{82.20}\\
 & w/o DS & 26.11 & 61.95 & 59.91 & 90.16 & \textbf{74.66} & 96.20 & \textbf{46.78} & 74.41 & 51.86 & 80.68\\
\bottomrule
\end{tabular}%
}
\caption{Ablation Study Results on Dynamic Rollout Budget Allocation.}
\label{tab:ablation}
\end{table*}

\begin{figure*}[!h]
    \centering 
    
    \begin{minipage}[!t]{0.45\textwidth}
        \centering 
        \includegraphics[width=\linewidth]{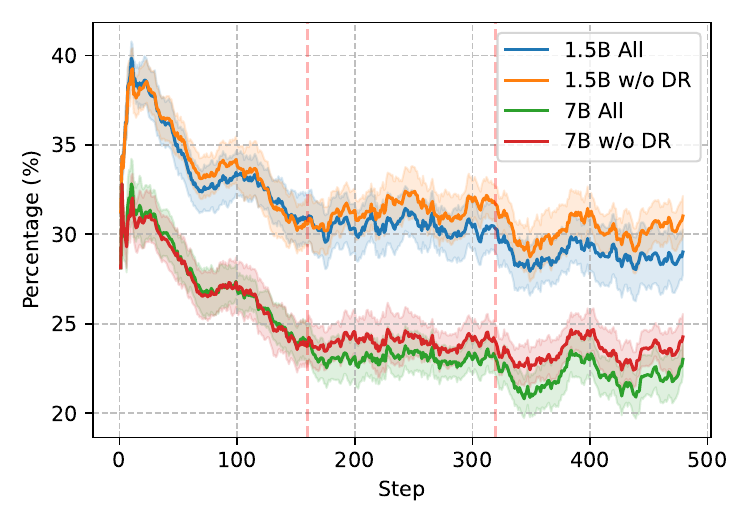}
        \vspace{-5pt} 
        \caption{The variation in the proportion of questions for which all rollouts are incorrect (smoothed) during the training process. The red vertical lines indicate the intervals between different data iteration rounds, which also correspond to the points where \( G_{\text{min}} \) and \( G_{\text{max}} \) are adjusted.}
        \label{fig:erro_rate}
    \end{minipage}
    \hfill 
    \begin{minipage}[!t]{0.45\textwidth}
        \centering 
        \includegraphics[width=\linewidth]{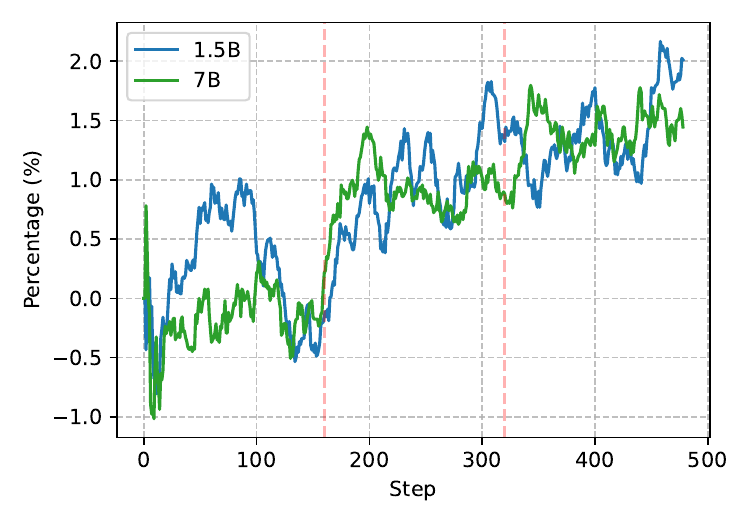}
        \vspace{-5pt} 
        \caption{The difference in the proportion of questions for which all rollouts are incorrect (smoothed) between using dynamic rollout budgeting (DR) and not using DR during the training process.}
        \label{fig:rate_dif}
    \end{minipage}
\end{figure*}

\noindent \textbf{What is the Role of Annealing?}
Figure \ref{fig:amc_passk} shows that annealing achieves greater improvements than temperature scheduling alone at lower $k$-values. However, as $k$ increases, models with annealing are gradually surpassed by those using only temperature scheduling. A similar trend is observed in Table \ref{tab:ts_cmp} on MATH 500 and Olympiad-Bench, where annealing outperforms at pass@1 but is overtaken by temperature scheduling alone at pass@16. 
On the more challenging AIME dataset, the temperature scheduler-only model consistently outperforms across all $k$-values, with the performance gap narrowing only at sufficiently high $k$. We attribute this phenomenon to this trade-offs: \textbf{Annealing improves precision on simpler questions by limiting the search space to high-value actions. In contrast, models trained exclusively with the temperature scheduler preserve a higher degree of exploratory capacity, facilitating the discovery of solution pathways for more complex questions.}

\subsection{Ablation Study on Dynamic Rollout Budget Allocation}

In the experiments conducted in this section, we investigate the impact of dynamic rollout budget allocation. 

\noindent \textbf{The Impact of Dynamic Rollout Budget Allocation on Performance.} As shown in Table \ref{tab:ablation}, the performance of the model deteriorates significantly on the most challenging AIME benchmark when dynamic rollout budgeting is not employed. The pass@1 scores on the AIME benchmark decreased by 3.02\% and 1.59\% for the 7B and 1.5B models, respectively. On other benchmarks, performance generally declines when dynamic rollout budget allocation is not used, with only a few cases showing slight improvements.

\noindent \textbf{The Impact of Dynamic Rollout Budget Allocation on the Proportion of Questions with All Incorrect Rollouts.} Figure \ref{fig:erro_rate} illustrates the proportion of questions for which all rollouts are incorrect during the training process. During the first data iteration, dynamic rollout budget allocation is not yet activated, resulting in a similar proportion of entirely incorrect rollouts with or without dynamic budget allocation.
During the second and third iterations, employing dynamic rollout budget allocation leads to a reduction in the proportion of questions for which all rollouts are incorrect.
As shown in Figure \ref{fig:rate_dif}, increase in \( G_{\text{max}} \) corresponds to a further reduction in the proportion of entirely incorrect rollouts. In the third iteration, when \( G_{\text{max}} \) is increased by 4 compared to \( G \), a reduction of approximately 1.5\% to 2\% in the proportion of entirely incorrect rollouts is achieved.


\section{Conclusion}


In this paper, we propose a dynamic rollout budget allocation mechanism to enhance the efficiency of reinforcement learning and a temperature scheduler to encourage greater exploration by the model. We conduct experiments on models with 1.5B and 7B parameters. Experimental results demonstrate that our method significantly outperforms GRPO training on the most challenging AIME 2024 benchmark. Additionally, on other benchmarks, models trained with our method achieve substantially higher pass@16 scores compared to those trained with GRPO. This indicates that models trained using our approach retain exploratory capabilities, enabling them to uncover more potential correct paths.

\section*{Limitations}

For the temperature scheduler, we have observed that annealing (low entropy) is more beneficial for simpler problems, while not using annealing (maintaining high entropy) is more advantageous for more challenging problems. Furthermore, we have modeled problem difficulty using the cumulative average reward. A natural idea, therefore, is to consider setting different temperatures for problems of varying difficulty. However, we have not conducted further experiments to explore this idea, leaving it as a direction for future work.

Due to computational resource constraints, we set the number of rollouts \( G \) per question to 8. However, increasing \( G \) and \( G_{\text{max}} \) could potentially amplify the effectiveness of dynamic rollout budget allocation.

Finally, although our approach can be easily extended to a broader range of reinforcement learning algorithms and domains, due to computational resource constraints, we limited our experiments to GRPO and the mathematics domain. Nevertheless, we believe that our method is sufficiently general and has the potential to be applied to other algorithms and domains.

\bibliography{arxiv.bib}

\appendix

\section{Entropy Scaling through Temperature Adjustment in Softmax Distributions}
\label{sct:entropy_scaling}

In this section, we address the adjustment of the temperature \(\tau\) such that the entropy is scaled by a factor of \(\alpha\). For simplicity, we focus on analyzing the distribution for the generation of a single next token. Let \(\mathbf{z} = (z_1, z_2, \dots, z_N)\) represent the logits generated by a LLM, and define the temperature as \(\tau > 0\). The commonly used Softmax probability distribution is defined as:
$$\mathbf{p}=\left[ p_1,p_2,\dots,p_N \right],$$
$$
p_i 
= \frac{e^{\,z_i/\tau}}{\sum_{j=1}^N e^{\,z_j/\tau}}.$$

Let \( z_\text{max} \) be the maximum value in the logits, and define \( \Delta_i = z_\text{max} - z_i \). Then, \( p_i \) can be expressed as:

\begin{align*}
    p_i 
&= \frac{e^{\,-(z_\text{max}-z_i)/\tau}}{\sum_{j=1}^N e^{\,-(z_\text{max}-z_j)/T}} \\
&= \frac{e^{-\Delta_i/T}}{\sum_{j=1}^N e^{-\Delta_j/\tau}} \\
&= \frac{e^{\,-\beta \Delta_i}}{Z(\beta)},
\end{align*}

where \( \beta = 1/\tau \) and \( Z(\beta) = \sum_{j=1}^N e^{-\beta \Delta_j} \) .The Shannon entropy \cite{shannon1948mathematical}  of this distribution is defined as:

\begin{equation}
H(\mathbf{p})
= -\sum_{i=1}^N p_i \,\ln p_i.    
\label{eq:entropy}
\end{equation}

Substituting $\ln p_i =- \beta \Delta_i - \ln Z(\beta)$ into Equation (\ref{eq:entropy}) results in the following decomposition:

\begin{align*}
H(\mathbf{p})
= &-\sum_{i=1}^N p_i\bigl(\,-\beta \Delta_i - \ln Z(\beta)\bigr)\\
= &\beta \sum_{i=1}^N p_i \Delta_i \;+\; \ln Z(\beta)\,\sum_{i=1}^N p_i. \\  
= &\beta\sum_{i=1}^N p_i\,\Delta_i \;+\; \ln Z(\beta).
\end{align*}

When the entropy \( H(\mathbf{p}) \) is small, the probability distribution \( \mathbf{p} \) is primarily concentrated on a specific state, with the contributions from other states being negligible. \citet{tang2024top} analyzed the distribution pattern of logits from LLMs and observed that they typically consist of a Gaussian-distributed noisy region and a distinct informative region containing a few outlier tokens. For simplicity, the analysis focuses on the logits associated with the most informative token, specifically considering only $z_\text{max}$. $z_\text{max}$ needs to be significantly larger than the logits in the noisy region to achieve a low entropy. We further assume that the difference between $z_\text{max}$ and the logits in the noisy region is approximately equal, i.e., $\Delta_j \approx \Delta$ for $z_j < z_\text{max}$. Under this assumption, the normalization factor of the distribution can be expressed as:

\[
Z(\beta) \approx 1 + (N-1)e^{-\beta\Delta}.
\]

When the entropy is small, the probability corresponding to \(z_\text{max}\) is given by \(p_\text{max} = \frac{1}{Z(\beta)} \approx 1\), which implies that \(Z(\beta) \approx 1\). Consequently, for the remaining \(N-1\) states, the probabilities can be approximated as:

\[
p_j = \frac{e^{-\beta\Delta}}{Z(\beta)} \approx e^{-\beta\Delta}.
\]

The entropy of the distribution can then be approximated as:

\begin{align*}
    H(p) =& -\sum_i^N p_i \ln p_i \\
    \approx &  -(N-1)e^{-\beta\Delta} \ln (e^{-\beta\Delta})\\
    = & (N-1)\beta\Delta e^{-\beta \Delta}
    .
\end{align*}

Suppose the initial entropy is approximately represented as \( \tilde{H}_0 =(N-1) \beta_0 \,\Delta \,e^{-\beta_0 \,\Delta} \). We aim to scale the entropy by adjusting the temperature \( \tau \), which is equivalent to modifying \( \beta \). Suppose we scale this entropy by \( \alpha \) times by adjusting \( \beta_0 \) to \( \beta_1 \), i.e., 
\begin{equation}
\alpha \,(N-1) \beta_0 \,\Delta \,e^{-\beta_0 \,\Delta} = (N-1)\beta_1 \,\Delta \,e^{-\beta_1 \,\Delta}.
\label{eq:a1}
\end{equation}

Assuming $\beta_1\Delta = \beta_0\Delta + d$, the ratio becomes:
\begin{align}
    \alpha=&\frac{(N-1)\beta_1\Delta e^{-\beta_1\Delta}}{(N-1)\beta_0\Delta e^{-\beta_0\Delta}} \nonumber\\
= &e^{-(\beta_1\Delta-\beta_0\Delta)}\,\frac{\beta_1\Delta}{\beta_0\Delta}\nonumber\\
=& e^{d}\frac{\beta_0\Delta+d}{\beta_0\Delta}
\label{eq:ratio}
\end{align}

The change in entropy introduced by a single training step is typically minimal, meaning that the scaling factor \( \alpha \) we aim to achieve is close to 1. we can further assume \( d \approx 0 \), allowing the approximation \( \beta_0\Delta + d \approx \beta_0\Delta \), Equation (\ref{eq:ratio}) can be approximately expressed as:
$\alpha \approx e^{-d}$. Then, by taking the natural logarithm, we obtain:

\begin{equation}
    d\approx -\ln \alpha.
\end{equation}

Thus, the ratio of the new temperature to the original temperature is:

\begin{align}
    \frac{\tau_1}{\tau_0}=& \frac{\beta_0}{\beta_1}=\frac{\beta_0\Delta}{\beta_1\Delta}\nonumber\\
\approx&
\frac{\beta_0\Delta}{\,\beta_0\Delta - \ln\alpha\,}.
\end{align}

When the entropy is low, \(\beta_0\Delta\) tends to be relatively large, while \(\ln \alpha\) is close to zero. Thus, it follows that \(\beta_0\Delta \gg \ln \alpha\). So, we can further approximate:

\begin{align}
    \frac{\tau_1}{\tau_0}
&\approx
\frac{\beta_0\Delta}{\,\beta_0\Delta-\ln \alpha\,}\\
&=\frac{\beta_0\Delta-\ln\alpha+\ln\alpha}{\,\beta_0\Delta-\ln \alpha\,}
= 
1 + \frac{\ln \alpha}{\,\beta_0 \,\Delta\,-\ln\alpha}\nonumber\\
&\approx 1+\frac{\ln\alpha}{\beta_0\Delta}.
\end{align}

Therefore, the new temperature can be expressed as:

\begin{equation}
    \tau_1 \approx \tau_0 \times \left(1 + \frac{\ln \alpha}{\beta_0\Delta}\right)=
\tau_0 \times \left(1 + \frac{\tau_0\ln \alpha}{\Delta}\right).
\label{eq:tp_relation_1}
\end{equation}

Equation (\ref{eq:tp_relation_1}) describes the approximate relationship between temperatures before and after adjustment. 

During training, logits can be utilized to estimate the value of \(\Delta\).
However, computing \(\Delta\) using logits incurs additional computational overhead and significant memory consumption, as the logits matrix is exceedingly large. To further simplify the computation, we introduce an approximation of the relationship between \(\Delta\) and \(N\), thereby eliminating the necessity of explicitly computing \(\Delta\) during training. In this context, we disregard the effect of temperature, and the entropy is assumed to be a small value, \(\varepsilon\). Based on the Equation (\ref{eq:tp_relation_1}), the entropy can be approximately expressed as:

\[
\varepsilon \approx (N-1)\Delta e^{-\Delta}.
\]

Taking the logarithm, we have:

\[
\ln(\varepsilon) \approx \ln(N-1) + \ln(\Delta) - \Delta.
\]

Rearranging terms, we obtain:

\[
\Delta \approx \ln(N-1) + \ln(\Delta) - \ln(\varepsilon).
\]

For sufficiently large \( N \), \( \ln(N-1) \) can be approximated as \( \ln(N) \), leading to:

\begin{equation}
    \Delta \approx \ln(N) + \ln(\Delta) - \ln(\varepsilon).
    \label{eq:delta}
\end{equation}

In Equation (\ref{eq:delta}), the dominant term on the right-hand side is \( \ln(N) \), while the other terms are comparatively smaller. Assuming \( \Delta = \ln(N) + c \), we substitute this expression into Equation (\ref{eq:delta}), yielding:

\begin{equation}
    \ln(N) + c \approx \ln(N) + \ln(\ln(N) + c) - \ln(\varepsilon).
    \label{eq:delta1}
\end{equation}

Simplifying Equation (\ref{eq:delta1}), we find:

\[
c \approx \ln(\ln(N) + c) - \ln(\varepsilon).
\]

For large \( N \), the value of \( c \) is expected to be much smaller than \(\ln(N)\). Thus, the addition of \( c \) to \(\ln(N)\) does not significantly change the logarithm. For the distribution of the next token generated from LLMs, the magnitude of \(\varepsilon\) is typically on the order of \(10^{-1}\). Thus, we can further neglect the \(\ln(\varepsilon)\) term, simplifying the expression. So \( c \) can be approximated as:

\[
c \approx \ln(\ln(N)).
\]

Therefore, \(\Delta \approx \ln(N) + \ln(\ln(N))\). Substituting this approximation for \(\Delta\) to Equation (\ref{eq:tp_relation_1}), the temperature scaling formula becomes:

$$\tau_1 \approx \tau_0 \times \left(1 + \frac{\tau_0 \ln \alpha}{\ln N + \ln\left(\ln N\right)}\right).$$

This provides an approximate formula for how the temperature needs to be adjusted to scale the entropy by a factor of $\alpha$.

\section{Dataset details}
\label{sct:dataset}
\begin{figure}[!h]
    \centering 
    \begin{minipage}[!t]{0.4\textwidth}
        \begin{minipage}[!t]{\textwidth}
        \centering 
        \includegraphics[width=\linewidth]{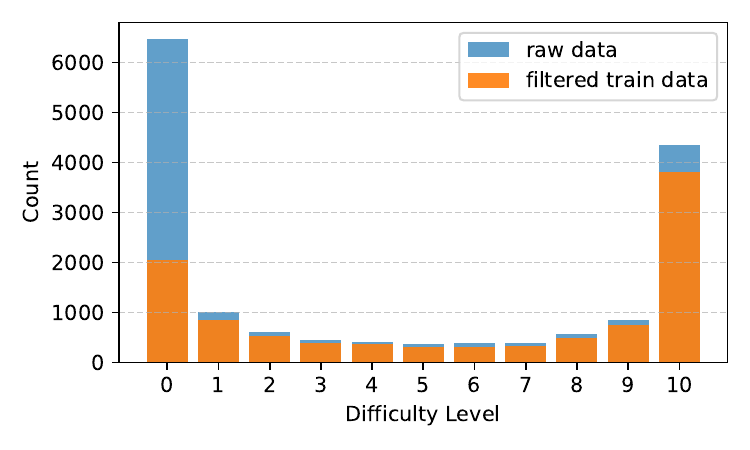}
        \vspace{-5pt} 
        \caption{Difficulty distribution of the validation set. The orange color is the difficulty distribution of the filtered 10k data, and the blue color is the difficulty distribution of the original data.}
        \label{fig:train_data_df}
    \end{minipage}
    \end{minipage}
    \hfill 
    \begin{minipage}[!t]{0.4\textwidth}
        \centering 
        \includegraphics[width=\linewidth]{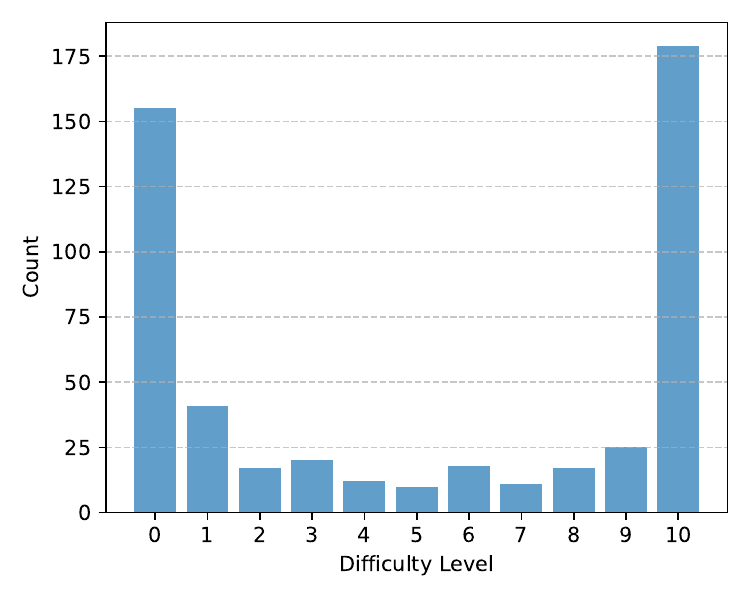}
        \vspace{-5pt} 
        \caption{Difficulty distribution of validation set. The difficulty distribution of the validation set is consistent with the original data.}
        \label{fig:valid_data_df}
    \end{minipage}
\end{figure}
\begin{figure}[!h] %
    \centering
    \includegraphics[width=0.85\linewidth]{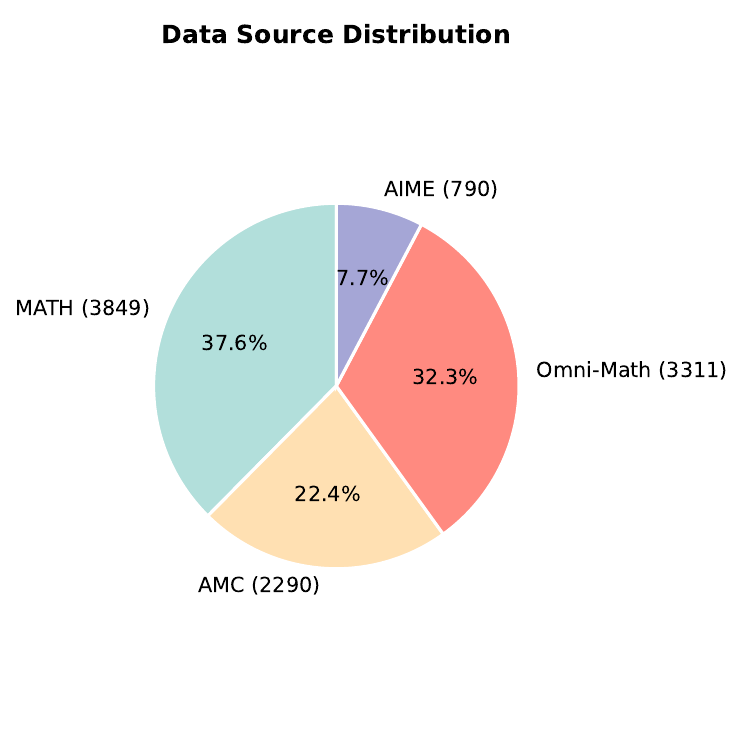} 
    \caption{Pie chart of data sources.}
    \label{fig:data_dist}
\end{figure}

We further processed the DeepScaleR 40k dataset \cite{deepscaler2025} to obtain our training data. Specifically, following the approach of Kimi k1.5 \cite{team2025kimi}, we improved the quality of reinforcement learning training data by reducing the proportion of simple questions.

We first utilized Qwen 2.5 Math 7B \cite{yang2024qwen2} to sample answers for each questions 10 times. The difficulty of the questions was assessed based on their accuracy rates. Subsequently, we balanced the data based on difficulty and filtered out a portion of simpler problems, capping the number of questions with 100\% accuracy at 2k. This balancing process not only ensures that the model is exposed to a diverse range of problem difficulties but also improves training efficiency by reducing the redundancy of overly simple problems, as training on simple problems is likely to yield minimal gains. The final 10k dataset exhibits a difficulty distribution as shown in Figure~\ref{fig:train_data_df}. The distribution of data sources is presented in Figure~\ref{fig:data_dist}.

For the validation set, data was evenly sourced from the MATH, Omni-Math, AMC, and AIME datasets, with 128 samples from each dataset, resulting in a total of 512 samples. We did not balance the difficulty of the validation set, ensuring that its difficulty distribution closely resembles that of the original dataset, as shown in Figure~\ref{fig:valid_data_df}. Furthermore, the validation set does not overlap with the training set.

\end{document}